%% file: iclr2020_cv4aFinal.tex
\documentclass{article} 
\usepackage{iclr2020_conference,times}

\input{math_commands.tex}

\usepackage{hyperref}
\usepackage{url}
\usepackage{graphicx}

\title{Mobile-Based Deep Learning Models for \\Banana Diseases Detection}

\author{Sophia Sanga, Victor Mero \& Dina Machuve \\
School of Computational and Communication Sciences and Engineering\\
Nelson Mandela African Institution of Science and Technology\\
PO Box 447, Arusha , Tanzania \\
\texttt{\{sangas,merov,dina.machuve\}@nm-aist.ac.tz} 
\And
Davis Mwanganda \\
Parrot AI  \\
Dar es Salaam,Tanzania \\
\texttt{\{davisdavid179\}@gmail.com}
}

\iclrfinalcopy 
\begin{document}

\maketitle

\begin{abstract}
Smallholder farmers in Tanzania are challenged on the lack of tools for early detection
of banana diseases. This study aimed at developing a mobile application for early detection of
Fusarium wilt race 1 and black Sigatoka banana diseases using deep learning. We used a dataset of 3000 banana leaves images. We pre-trained our model on Resnet152 and Inceptionv3 Convolution Neural Network architectures. The Resnet152 achieved an accuracy of 99.2\% and Inceptionv3 an accuracy of 95.41\%. On deployment using Android mobile phones, we chose Inceptionv3 since it has lower memory requirements compared to Resnet152. The mobile application on real environment detected the two diseases with a confidence level of 99\% of the captured leaf area. This result indicates the potential in improving the yield of bananas by smallholder farmers using a tool for early detection of diseases.

\end{abstract}

\section{Introduction}

Bananas are a staple food for about 70 million people in Africa and are mainly produced by smallholder farmers \citep{etebu2011control}. The East Africa region (including Tanzania) is the largest producer with Uganda leading in the region and it is the second largest producer globally after India \citep{tripathi2009xanthomonas}. In Tanzania, bananas are mainly produced in Arusha, Mbeya and Kilimanjaro regions. However, the yield is highly affected by fungal diseases that include Fusarium wilt race 1 and black Sigatoka resulting into high losses in crop  \citep{deltour2017disease,chillet2009sigatoka,gutierrez2015effective}. In Tanzania, smallholder banana farmers in Arusha region acknowledge the presence of banana fungal diseases on their plantations. However, diseases detection has been a challenge to farmers since majority still rely on traditional knowledge \citep{ramadhani2017identification}. Crop management practices and information pertaining to crop diseases is mainly disseminated by agricultural extension officers and local plant clinics. A limited number of farmers have access to extension services. This is due to the few extension officers serving farmers at district level, for example in Arumeru district at Akheri ward, one extension officer serves up to 1,700 farmers \citep{ramadhani2017identification}.

Computer vision is a field of deep learning with applications in agriculture for crop diseases detection. Crop diseases diagnostics for cassava and bananas in East Africa is an application of computer vision in agriculture \citep{owomugisha2014automated,mwebaze2016machine}. The study conducted in Tanzania on the detection of cassava diseases used transfer learning on convolution neural network (CNN) framework to pre-train models on cassava leaves images. The models were deployed on mobile phones to detect brown leaf spot (BLS), cassava mosaic disease (CMD), red mite damage (RMD), cassava brown streak disease (CBSD) and green mite damage (GMD) diseases \citep{ramcharan2019mobile}. The motivation of our work was extending the study on cassava diseases detection to banana fungal diseases using banana leaf images.
The next sections cover the materials and methods used to train and test the selected models, the pre-processing of the collected images and model deployment on Android mobile phones. 


\section{MATERIALS AND METHODS}
\label{gen_inst}

\subsection{Deep Convolutional neural network models}
Deep Convolutional Neural Networks have been used in many filed such as computer vision, speech recognition, face recognition, and natural language processing. However in computer vision deep learning was found more effective for image recognition, Object Recognition, self-driving cars, robotic, Object Detection and Image Segmentation tasks \citep{voulodimos2018deep}. Architectures of Convolutional Neural Networks have been developed, which have been effectively applied to complicated tasks of visual imagery \citep{mohanty2016using}. 
The two basic Convolutional Neural Networks architectures that we tested in the problem explored for the early detection of banana diseases from images of their leaves, were the following: Resnet152 \citep{he2016deep}, InceptionV3 \citep{szegedy2017inception}. We trained the pre-trained  models using TensorFlow since it allows users to train models on both Graphics Processing Units (GPUs) and Central Processing Units (CPUs) \citep{abadi2016tensorflow}. We carried out the experiments in two different machines, for Resnet152 a laptop was used, with these specifications: window 10 Pro N equipped with one Processor- Intel(R)Core(TM)i5-4300U CPU @1.90GHz 2.50 GHz with RAM size of 4GB  and system type is 64-bit Operating system. Furthermore, the language used during training was python as it is a programming language. The implementation was carried out using Keras library for neural network models and has high performance for numerical computation with TensorFlow on the back-end. The software library used to train the models was the Google Colab machine with the following specification: run-time type is Python3 and hardware accelerator is GPU and Notebook size of 20Mb was used. We trained on the InceptionV3 using a Tower computer with the following specifications: Dell Precision T7610 Workstation, memory size of 256 GiB, processor Intel Xeon(R)CPUE5-26200@2.00GHz*12,Graphics NVD9,running Ubuntu 18.04 64bits and Hard disk 1.0TB.

\subsection{Training and testing datasets}

On data pre-processing for the 3000 images from each class, we used data augmentation techniques for image data generator for horizontal flipping, zoom range of 0.2,rotating-right-bottom, rotating-left-bottom, shear range of 0.1, cropping and Re-scaling each image with the ration of 1./255 factor. Our final dataset size was 18,000 banana leaf images of three class Black Sigatoka, Fusarium Wilt race 1 and healthy leaves as shown in Figure 1. Data analysis and pre-processing steps that we have taken in model development. In order to avoid bias in our pre-trained models, the data was randomly shuffled before splitting. Image dataset was then split in the following ratio: 80\% training, 15\% validation, 5\% testing. The training and validation set was used during model selection while the test set was used to evaluate the generalization performance (performance verification stage) of the selected model.

\section{RESULTS}

The selected models above were trained and their results are presented in Table \ref{perfomance} after several experiments we select these values because they show the best results during training.
The two models achieve better performance when tested by using original banana leaves
images.

\begin{table}[]
	\caption{ Performance of the models architectures for detection of banana disease
	}
	\label{perfomance}
	\begin{center}
		\begin{tabular}{llllll}
			\hline 
			\multicolumn{1}{c}{\bf Models}  &\multicolumn{1}{c}{\bf Validation accuracy \%}  &\multicolumn{1}{c}{\bf Test accuracy \% }  &\multicolumn{1}{c}{\bf Epoch} &\multicolumn{1}{c}{\bf Time(s/epoch)} &\multicolumn{1}{c}{\bf Loss} \\ \hline
			Resnet152   & 0.992                                                             & 0.998                                                      & 50    & 515                                                       & 0.0539 \\ \hline
			InceptionV3 & 0.954                                                             & 0.955                                                      & 150   & 159                                                       & 0.1351 \\ \hline
		\end{tabular}
	\end{center}
\end{table}

\section{MOBILE DEPLOYMENT}
\label{others}

We choose InceptionV3 for mobile deployment because it takes less memory when deployed
to mobile apps as compared to other deep learning models \citep{meng2017training,szegedy2017inception}. However resnet152 archives higher accuracy when
compared to inception V3 but it requires large memory space for the mobile deployment. We use Tensor flow as it is a deep learning framework that offers APIs for mobile deployment
in Android and IOS based smartphones \citep{abadi2016tensorflow}. The tensor flow was used to train the InceptionV3 model for banana disease detection. Figure 1 show the block diagram of the banana disease detection mobile application deployment block diagram. The validation of the deployed model was done and screenshots of available in the appendix section of this paper.
\begin{figure}[h]
	\begin{center}
		\includegraphics[width=1.0\textwidth]{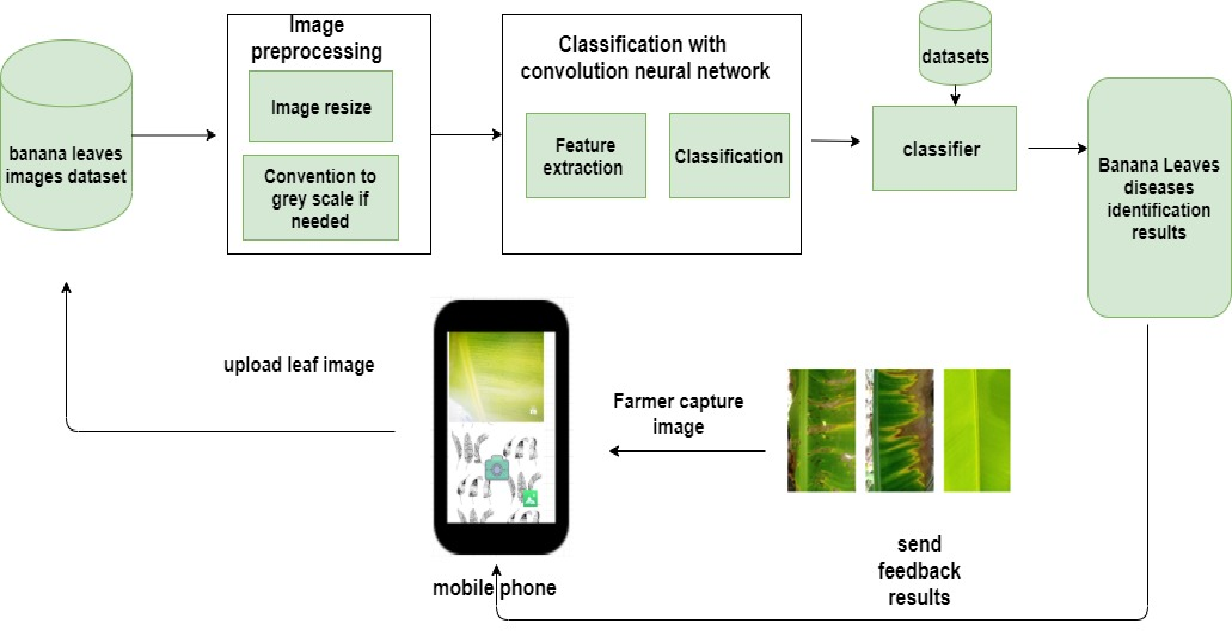}
	\end{center}
	\caption{Banana disease detection mobile application deployment block diagram.}
\end{figure}

\section{CONCLUSION}

In this work we evaluated the capability of deep learning
models and transfer learning techniques in crop diseases diagnostics. We deployed the model on mobile phone with capability to operate offline for the early detection of banana
diseases. We assessed the performance of the deployed tool to detects banana diseases from the
for the targeted three classes. In both tests we set our result to at least have
the percentage confidence level in detection to be at least 70\%, otherwise the app will recommend having a clear image of the captured leaf. This study concludes that early detection on real time is very important for banana diseases to improve banana yields, which are largely affected by
diseases commonly being black Sigatoka and Fusarium wilt race 1. The study shows that
with the use of machine learning techniques and mobile phones these diseases can be easily
detected and then controlled in proper ways. Our developed mobile application is called FUSI
Scanner \ (Fusarium Sigatoka) using android technology to detect the presence of two mainly
banana fungal diseases.

\bibliography{iclr2020_conference}
\bibliographystyle{iclr2020_conference}


\end{document}

%% file: math_commands.tex
\usepackage{amsmath,amsfonts,bm}

\def\eqref#1{equation~\ref{#1}}

\def\1{\bm{1}}

\DeclareMathAlphabet{\mathsfit}{\encodingdefault}{\sfdefault}{m}{sl}
\SetMathAlphabet{\mathsfit}{bold}{\encodingdefault}{\sfdefault}{bx}{n}

